\documentclass[10pt,twocolumn,letterpaper]{article}

\usepackage{cvpr}              

\usepackage{graphicx}
\usepackage{amsmath}
\usepackage{amssymb}
\usepackage{booktabs}
\usepackage{multirow}

\usepackage[pagebackref,breaklinks,colorlinks]{hyperref}

\usepackage[capitalize]{cleveref}
\crefname{section}{Sec.}{Secs.}
\Crefname{section}{Section}{Sections}
\Crefname{table}{Table}{Tables}
\crefname{table}{Tab.}{Tabs.}

\def\cvprPaperID{*****} 
\def\confName{CVPR}
\def\confYear{2026}

\begin{document}

\title{DROID-ANCHOR: Odometry-Anchored Recurrent Metric Depth Estimation}

\author{Yuxuan Chen\\
UC Berkeley\\
{\tt\small cyx1024@berkeley.edu}
\and
Brook Du\\
UC Berkeley\\
{\tt\small brookdu@berkeley.edu}
}
\maketitle

\begin{abstract}

Precise metric depth estimation is fundamental for autonomous robot navigation, yet monocular systems inherently suffer from scale ambiguity and scale drift. While recent recurrent flow-based SLAM systems have demonstrated state-of-the-art robustness, they remain scale-ambiguous. In this paper, we propose Metric-DROID, an end-to-end recurrent architecture that anchors visual SLAM to physical reality by integrating proprioceptive odometry. Our framework introduces the following innovations: (1) a LSTM Update Operator that encodes high-frequency odometry sequences into spatial feature maps, providing a persistent metric bias for iterative refinement. (2) an Uncertainty-Aware Metric Backend ($BA_{odom}$) that treats odometry as a geometric anchor with learned heteroscedastic covariance. By regressing a time-varying metric uncertainty $\Sigma_{o}$, our system intelligently balances visual re-projection and metric translation residuals, effectively mitigating the impact of wheel-slip and sensor noise. (3) We further propose a selective residual fine-tuning strategy to preserve pre-trained geometric priors while enabling zero-shot metric alignment. 

\end{abstract}

\section{Introduction}
\label{sec:introduction}

For mobile robots, standard navigation tasks such as obstacle avoidance and path planning require precise metric distances, making mere relative proximity information insufficient. While RGB-D and stereo cameras can provide accurate metrics, their high cost and calibration complexity remain prohibitive for mass-market robotic platforms. Consequently, achieving accurate metric perception using only a ubiquitous monocular camera would significantly democratize accessible and low-cost robotic navigation. Furthermore, acquiring this physical accuracy must not incur significant computational overhead, as low-latency inference is strictly required for real-time closed-loop control.

Despite the success of deep SLAM systems~\cite{teed2021droidslam, murai2024mast3r} in achieving relative consistency, a fundamental dilemma persists between metric accuracy and real-time inference. On one hand, state-of-the-art monocular models~\cite{lu2025align3r,wang2023dust3r,wang2025vggt, murai2024mast3r} remain scale-blind, suffering from inherent scale ambiguity and catastrophic drift over long-range trajectories. On the other hand, contemporary metric-aware pipeline MOGS~\cite{zhang2026mogs} leverage object features to anchor metric, however introduce excessive computational latency. This latency creates a temporal misalignment between the estimated spatial state and the robot’s immediate physical pose, rendering spatial information obsolete for high-frequency closed-loop control. While other solutions like MAC-VO~\cite{qiu2025mac} attempt to balance this, they still rely on the stereo camera. Consequently, a real-time system that simultaneously satisfies global metric grounding and low-latency feedback remains an open challenge.

Thus, we aim to bridge between precise metric depth and real-time efficiency by utilizing the robot’s
internal odometry (wheel encoders or IMU) as a persistent
scale constraint.

\begin{figure}[t]
  \centering
   \includegraphics[width=1.0\linewidth]{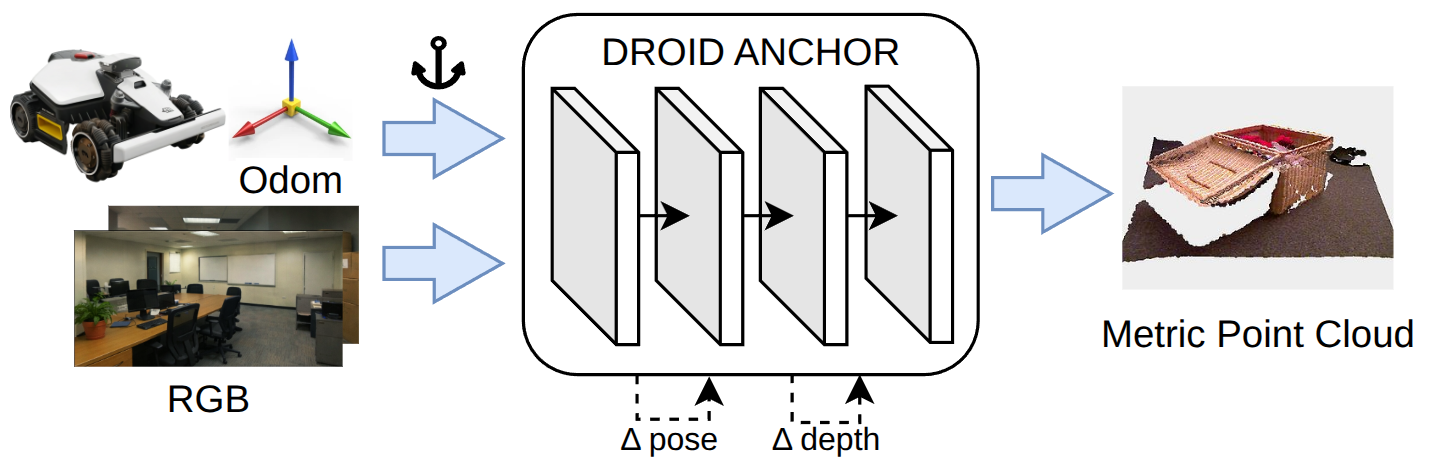}
   \caption{Our framework, DROID ANCHOR, utilizes raw robot odometry as a persistent geometric anchor to resolve monocular scale ambiguity}
   \label{fig:teaser}
\end{figure}

\section{Related Work}
\textbf{Deep Monocular Geometry.} Recent breakthroughs in feed-forward geometric models, such as DUSt3R~\cite{wang2023dust3r} and VGGT~\cite{wang2025vggt}, have redefined 3D reconstruction by estimating geometry directly from image pairs. While these methods produce high-fidelity point clouds, they primarily focus on visualization and rendering. As monocular systems, they lack an absolute reference, resulting in "virtual" scale units that cannot be directly utilized for robotic metric navigation. Our work differs by anchoring these geometric priors to physical units via a persistent odometry constraint.

\textbf{Recurrent Visual SLAM.} DROID-SLAM~\cite{teed2021droidslam} introduced an iterative refinement framework using ConvGRUs, achieving unprecedented robustness against rapid motion and low-texture environments. Subsequent variants like DROID-W~\cite{li2026droidslamwild} have further improved generalization in the dynamic environment. However, these systems operate on an relative scale. While they achieve excellent relative tracking, the lack of metric grounding leads to scale drift, which is a fatal flaw for real-world obstacle avoidance. We build upon the recurrent architecture of DROID-SLAM but reformulate the backend optimization to incorporate metric anchors.

\textbf{Metric Scale Recovery.} Recovering absolute scale from monocular sequences typically involves auxiliary cues, such as object categories (Object-SLAM) or multi-sensor fusion. Recent works like MOGS~\cite{zhang2026mogs} achieve metric scale but at the cost of high-latency multi-stage processing. MAC-VO~\cite{qiu2025mac} addresses efficiency by utilizing a probabilistic weighting mechanism for odometry. However, it requires stereo camera input and the frame-to-frame nature lacks the global geometric consistency of graph-based BA. Unlike these methods, DROID method integrates odometry as both a latent feature prior and a hard geometric constraint within a recurrent BA framework, achieving a superior trade-off between metric fidelity and real-time throughput.

\section{Methodology}

\begin{figure*}[t]
  \centering
   \includegraphics[width=0.65\linewidth]{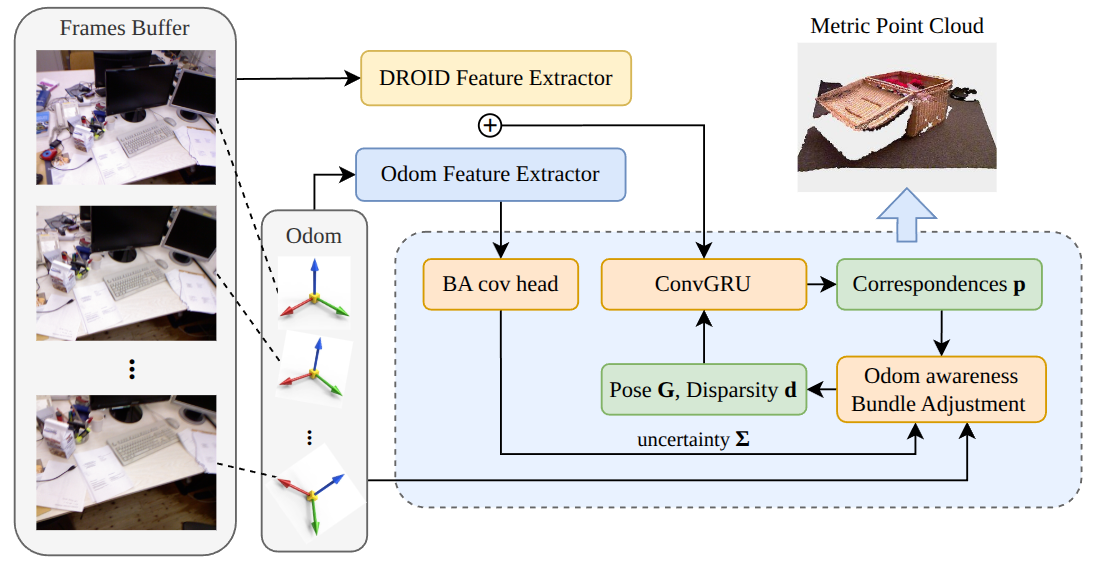}
   \caption{Overview of the Metric-DROID Architecture. Our framework extends the recurrent refinement paradigm with two metric-anchored components:
    (a) Odom Feature Extractor: Beyond visual correlation and motion features, we introduce a proprioceptive branch. An Odom-Sequence Encoder (LSTM) processes relative pose sequences, which are then broadcasted into spatial feature maps and fused into the ConvGRU hidden state.
    (b) Uncertainty-Aware Backend: An auxiliary Metric Covariance Head regresses the heteroscedastic uncertainty $\Sigma_{o}$ for each edge in the frame graph.
    (c) Metric Bundle Adjustment: The optimization layer treats odometry as a persistent geometric anchor, dynamically weighting metric residuals via $\Sigma_{o}$ to achieve deterministic physical scale while mitigating sensor drift and wheel-slip.}
   \label{fig:framework}
\end{figure*}

In this section, we present our Odometry-Anchored Recurrent framework, which synergizes high-frequency odometry priors with visual features to achieve robust, metric-scale scene understanding. Our approach integrates motion dynamics into the feature space and enforces global consistency through a uncertainty-aware bundle adjustment formulation

\subsection{Proprioceptive Feature Injection}
We process high-frequency odometry sequences through a two-stage hierarchical encoding: 

\textbf{Temporal Encoding:} 
Following the sequential learning framework in VINet~\cite{clark2017vinet}, we encode the temporal evolution of camera motion using an LSTM over a sequence of relative $\mathrm{SE}(3)$ transformations. 

We represent the absolute pose of frame $i$ in the world coordinate system as:
\begin{equation}
T_i = 
\begin{bmatrix}
R_i & t_i \\
0 & 1
\end{bmatrix}, \quad
T_{i+1} = 
\begin{bmatrix}
R_{i+1} & t_{i+1} \\
0 & 1
\end{bmatrix},
\end{equation}
where $T_i \in \mathrm{SE}(3)$ maps points from the camera coordinate frame at time $i$ to the world coordinate frame. The inverse transformation
\begin{equation}
T_i^{-1} =
\begin{bmatrix}
R_i^\top & -R_i^\top t_i \\
0 & 1
\end{bmatrix}
\end{equation}
maps world coordinates back to the local frame at time $i$.

The relative motion between consecutive frames is computed as:
\begin{equation}
\Delta T_{i \rightarrow i+1} = T_i^{-1} T_{i+1} =
\begin{bmatrix}
R_i^\top R_{i+1} & R_i^\top (t_{i+1} - t_i) \\
0 & 1
\end{bmatrix},
\end{equation}
which represents the motion from frame $i$ to $i+1$ expressed in the local coordinate system of frame $i$. We extract the relative translation and rotation as:
\begin{equation}
\Delta t_{i \rightarrow i+1} = R_i^\top (t_{i+1} - t_i), \quad
\Delta R_{i \rightarrow i+1} = R_i^\top R_{i+1}.
\end{equation}

The sequence of relative motions $\{\Delta T_{k \rightarrow k+1}\}$ between frames $i$ and $j$ is then fed into the LSTM encoder, producing a compact latent representation $\mathbf{z}_{ij}$ that captures the underlying motion dynamics.

\textbf{Spatial Broadcasting:} 
To integrate motion information with spatial visual features, we map the latent motion representation $\mathbf{z}_{ij}$ to a 2D feature map via an Odometry MLP. This spatially broadcasted representation is then concatenated with visual features (e.g., cost volumes and optical flow), providing an explicit metric prior to guide the ConvGRU during iterative hidden state refinement. This fusion enables the network to incorporate both geometric motion cues and image-based correspondences for accurate metric reasoning.

\subsection{Bundle Adjustment with Odometry}
To ensure global geometric consistency and resolve the inherent scale ambiguity in monocular SLAM, we formulate the SLAM backend as a joint optimization of visual re-projection and metric translation residuals. 

The total energy function is defined as:

\begin{align}
\begin{split}
E(\mathbf{G}, &\mathbf{d}) = \sum_{(i,j) \in \mathcal{E}} \| \mathbf{p}^*_{ij} - \Pi_c(\mathbf{G}_{ij} \circ \Pi_c^{-1}(\mathbf{p}_i, d_i)) \|^2_{\Sigma_{ij}}\\
&+ \lambda_{odom} \sum_{(i,j) \in \mathcal{E}} \| \Delta\mathbf{G}_{ij} - \Delta \mathbf{O}_{ij} \|^2_{\Sigma_{odom}}
\end{split}
\end{align} 
where $\mathbf{r}_{vis}$ denotes the photometric re-projection error, and $\mathbf{r}_{odom} = \text{Trans}(\mathbf{G}_j \mathbf{G}_i^{-1}) - \Delta \mathbf{O}_{ij}$ represents the metric displacement residual. This formulation anchors the visual scale to the physical manifold defined by the robot’s proprioceptive sensors. Furthermore, inspired by the probabilistic weighting mechanisms in MAC-VO~\cite{qiu2025mac}, we substitute static hyper-parameters with a learnable Metric Covariance Head. This module regresses a heteroscedastic covariance $\Sigma_{odom, ij}$ that captures the non-Gaussian noise characteristics inherent in robotic locomotion. By dynamically adjusting the precision matrix, the system effectively mitigates catastrophic scale drift caused by wheel-slip or sensor degradation.

Furthermore, to solve the resulting non-linear least squares problem via the Gauss-Newton algorithm, we perform incremental updates on the Lie algebra $\mathfrak{se}(3)$ using the right perturbation formulation. Specifically, let the translation residual be $\mathbf{r}_o = \text{trans}(\mathbf{G}_j \mathbf{G}_i^{-1}) - \mathbf{o}_e$. We derive the analytical Jacobians of $\mathbf{r}_o$ with respect to the camera poses $\mathbf{T}_i$ and $\mathbf{T}_j$ as $\mathbf{J}^o_j = [\mathbf{I}_{3 \times 3} \mid -[\mathbf{t}_{ij}]_\times]$ and $\mathbf{J}^o_i = -\mathbf{J}^o_j \cdot \text{Ad}(\mathbf{G}_{ij})$, respectively. These Jacobians enable the seamless integration of metric constraints into the sparse Hessian matrix. By solving the reduced camera system via the Schur complement, our framework jointly optimizes the global pose graph and the metric scale of the estimated depth maps, ensuring geometric consistency across the sequence. The detailed 
mathematical derivation is given in the appendix~\ref{app:derivation} part. 

\subsection{Noise Modeling }
In real-world robotic applications, obtaining synchronized, high-quality RGB-D and odometry datasets is often labor-intensive and platform-specific. Furthermore, odometry characteristics vary significantly across different robotic platforms due to disparate sensor modalities, varying wheel diameters, and distinct mounting extrinsics. To ensure that our model generalizes across heterogeneous hardware without requiring extensive site-specific recalibration, we propose a synthetic noise modeling framework to simulate these diverse sensor data during training.

We model the corrupted odometry observation $\mathbf{o}_e$ by sequentially applying scale and additive perturbations to the ground-truth translation $\mathbf{t}^{gt}$:
\begin{align}
\mathbf{s}_e &= 1 + \epsilon^s, \label{eq:scale_noise} \\
\mathbf{o}_e &= s_e \cdot \mathbf{t}^{gt} + \epsilon^t, \label{eq:additive_noise}
\end{align}
where $\mathbf{s}_e$ denotes the scale factor used to simulate systematic biases (e.g., wheel-slip), and $\epsilon^t$ represents additive Gaussian noise capturing high-frequency jitter during robotic locomotion. Here, $\epsilon^s$ and $\epsilon^t$ are sampled from zero-mean Gaussian distributions with task-specific variances.

\subsection{Training Objective}

We optimize a multi-term objective that enforces consistency across geometry, reprojection, pixel motion, depth, and odometry:
\begin{equation}
\mathcal{L}_{total} = 
w_1 \mathcal{L}_{disp}
+ w_2 \mathcal{L}_{odom}.
\end{equation}
To account for iterative refinement, all loss terms are weighted across optimization steps:
\begin{equation}
\alpha_k = \gamma^{(n-k-1)}, \quad \gamma \in (0,1),
\end{equation}
where $k$ indexes the refinement step and later predictions receive higher weights.

\textbf{Depth Loss.}
We supervise the predicted depth using valid ground-truth measurements:
\begin{equation}
\mathcal{L}_{disp} =
\sum_k \alpha_k \,
\frac{
\sum_{\mathbf{x}} m(\mathbf{x}) 
\left| d^{(k)}(\mathbf{x}) - d_{gt}(\mathbf{x}) \right|
}{
\sum_{\mathbf{x}} m(\mathbf{x})
},
\end{equation}
\begin{equation}
m(\mathbf{x}) =
\begin{cases}
1, & d_{gt}(\mathbf{x}) > 0 \\
0, & \text{otherwise}
\end{cases}
\end{equation}
Rather than treating all pixels equally, we restrict supervision to valid depth regions, ensuring stable training in the presence of missing measurements.

\textbf{Odometry Likelihood Loss.}
To model uncertainty in odometry observations, we adopt a heteroscedastic Gaussian likelihood formulation following~\cite{kendall2017uncertainties}:
\begin{equation}
r = \Delta O_{obs} - \Delta O_{gt},
\end{equation}
\begin{equation}
\mathcal{L}_{odom} =
\sum_k \alpha_k \, \mathbb{E} \left[
\frac{1}{2} \left( \frac{r}{\sigma^{(k)}} \right)^2 
+ \log \sigma^{(k)}
\right],
\end{equation}
where $\sigma^{(k)}$ is a learnable uncertainty parameter.
This formulation enables the model to down-weight unreliable odometry signals while emphasizing more confident estimates, leading to improved robustness under noisy or inconsistent motion measurements.

Taken together, these objectives jointly enforce consistency across geometric motion, reprojection, pixel correspondences, depth estimation, and odometry priors, leading to robust and metrically grounded scene understanding.

\subsection{Selective Residual Fine-tuning}
To facilitate efficient training while preserving the robust geometric priors of the original framework, we implement a selective freezing strategy during the fine-tuning stage.

\textbf{Parameter Preservation.} We freeze the weights of the feature extraction network ($f_{net}$) and the context network ($c_{net}$) from the pre-trained original DROID-SLAM.

\textbf{Gradient-Constrained GRU Update.} To incorporate the newly introduced metric bias, we expand the input dimension of the ConvGRU from 448 to 512. To prevent the degradation of pre-established geometric reasoning, we employ gradient hooks to implement a partial update strategy. Specifically, we only permit gradients to backpropagate through the weights associated with the newly added 64-channel metric features, while keeping the weight slices corresponding to the original 448-dimensional visual features fixed. This surgical update ensures that the network learns to fuse metric priors without destabilizing the pre-trained iterative hidden state refinement.

\section{Experiments}

\subsection{Implementation and Experimental Setup}

\textbf{Implementation}. All experiments run on a single NVIDIA
GeForce RTX 4080 laptop GPU and an Intel Core i9-13950HX CPU (5.50 GHz). 

\textbf{Datasets and Metrics.}
We evaluate our method on the TUM RGB-D benchmark~\cite{sturm12iros}, which provides monocular RGB images, depth maps, and ground-truth camera trajectories with metric scale. 

We report performance in terms of metric depth accuracy and computational efficiency.
For depth accuracy, we use standard depth estimation metrics, including Absolute Relative Error (AbsRel), Root Mean Squared Error (RMSE), and threshold accuracy $\delta_1$, defined as the percentage of pixels satisfying 
$\max(d^*/d, d/d^*) < 1.25$, 
where $d$ and $d^*$ denote the predicted and ground-truth depth values, respectively.

\subsection{Performance Evaluation}
\textbf{Depth Accuracy.}
We compare our method against DROID-SLAM as representative visual SLAM baselines. We attempted to include MOGS \cite{zhang2026mogs}; however, the open-source package is currently not complete.
In addition, we include Depth Anything V3~\cite{depthanything32025lin} as a recent method capable of recovering metric-scale geometry, providing a strong reference for evaluating absolute depth accuracy. 
These comparisons allow us to assess both geometric performance and practical efficiency across different approaches.

While Depth Anything V3 achieves superior metrics than Droid-Anchor, it functions as a pre-trained monocular depth estimation model leveraging massive data priors. In contrast, Droid-Anchor is a complete visual SLAM system designed for real-time camera tracking and consistent map reconstruction.

\begin{table}[h]
    \centering
    \renewcommand{\arraystretch}{1.2}
    \begin{tabular}{lccc}
        \toprule
        & \multicolumn{3}{c}{TUM-Dataset} \\
        \cmidrule(lr){2-4}
        Method & AbsRel$\downarrow$ & RMSE$\downarrow$ & $\delta_1 \uparrow$ \\
        \midrule
        Depth Anything V3 & 0.136 & 0.342 & 0.658 \\
        Droid-SLAM & 0.617 & 6.800 & 0.373 \\
        \textbf{Droid-Anchor (Ours)} & 0.528 & 1.663 & 0.392 \\
        \bottomrule
    \end{tabular}
\end{table}

\subsection{Ablation Study}

We conduct two ablation studies to evaluate the contribution of temporal odometry modeling and odometry noise modeling. First, to assess whether temporal modeling is beneficial for metric-scale depth estimation, we replace the LSTM-based odometry encoder with a simpler mean-pooling encoder. In the baseline setting, the relative odometry sequence is encoded by an LSTM before being projected into the odometry feature map. Both models use the same odometry bundle adjustment setting.

\begin{table}[h]
    \centering
    \renewcommand{\arraystretch}{1.2}
    \begin{tabular}{lccc}
        \toprule
        & \multicolumn{3}{c}{TUM-Dataset} \\
        \cmidrule(lr){2-4}
        Method & AbsRel$\downarrow$ & RMSE$\downarrow$ & $\delta_1 \uparrow$ \\
        \midrule
        Mean Pooling Ablation & 1.644 & 7.658 & 0.234 \\
        \textbf{Droid-Anchor (Ours) } & \textbf{0.528} & \textbf{1.663} & \textbf{0.392} \\
        \bottomrule
    \end{tabular}
    \caption{Ablation study on the odometry encoder. The LSTM-based encoder achieves better overall performance than the mean-pooling ablation, suggesting that temporal odometry modeling provides a more informative motion prior for metric depth estimation.}
    \label{tab:lstm_ablation}
\end{table}

Second, we evaluate the effect of the proposed odometry noise modeling strategy. The baseline model is trained with randomly perturbed odometry inputs, including translation, rotation, and scale perturbations, while the ablation model removes this noise modeling and is trained directly with clean GT-derived odometry. 

\begin{table}[h]
    \centering
    \renewcommand{\arraystretch}{1.2}
    \begin{tabular}{lccc}
        \toprule
        & \multicolumn{3}{c}{TUM-Dataset} \\
        \cmidrule(lr){2-4}
        Method & AbsRel$\downarrow$ & RMSE$\downarrow$ & $\delta_1 \uparrow$ \\
        \midrule
        Clean Odom Ablation & 1.122 & 5.619 & 0.260 \\
        \textbf{Droid-Anchor (Ours) } & \textbf{0.528} & \textbf{1.663} & \textbf{0.392} \\
        \bottomrule
    \end{tabular}
    \caption{Ablation study on odometry noise modeling. Training with synthetic odometry perturbations improves AbsRel and RMSE compared with directly using clean GT-derived odometry, indicating that noise modeling enhances robustness to imperfect odometry inputs.}
    \label{tab:noise_ablation}
\end{table}

As shown in Table~\ref{tab:lstm_ablation}, replacing the LSTM encoder with mean pooling leads to inferior overall performance. This indicates that temporal odometry sequences contain structured motion cues that cannot be fully captured by order-invariant aggregation. In contrast, the LSTM encoder can model and integrate more complex frame-to-frame motion variations, providing a more informative motion prior for metric-scale depth estimation. Table~\ref{tab:noise_ablation} further shows that removing odometry noise modeling degrades AbsRel and RMSE. The synthetic odometry perturbations therefore serve as an effective regularization strategy by exposing the model to translation, rotation, and scale variations during training, improving robustness to odometry noise and camera-odometry extrinsic misalignment across different camera configurations.

\section{Conclusion \& Discussion}
In this paper, we introduced DROID-ANCHOR, a novel recurrent framework designed to resolve the inherent scale ambiguity in monocular SLAM by anchoring visual geometry to proprioceptive odometry. By tightly coupling metric priors within the recurrent update operator, our model demonstrates a robust capability for deterministic metric scale estimation.

Despite these advancements, we acknowledge that the generalization of the current model is constrained by the scarcity of large-scale datasets that simultaneously provide high-fidelity RGB-D ground truth and synchronized odometry. Furthermore, the inherent variance in the odometry noise profiles on different robotic platforms poses a challenge for generalization of the cross-data set. To address these limitations and further validate the system’s robustness, our future work will focus on the collection of a comprehensive real-world dataset. This will involve deploying a mobile robot in high-traffic indoor environments to evaluate the performance of the framework against dynamic occlusions, complex motion patterns, and various mechanical noise.

\newpage
{\small
\bibliographystyle{ieee_fullname}
\bibliography{egbib}
}

\newpage
\appendix
\section{BA Mathematical Derivations} \label{app:derivation}

In this section, we provide a detailed derivation of the Jacobian matrices and the construction of the normal equations for the proposed Odometry-Anchored Bundle Adjustment ($BA_{odom}$).

\subsection{Objective Function}

The total energy functional $E$ is composed of a visual re-projection term and a metric odometry term: 

\begin{equation}
\begin{aligned} &E(\mathbf{G}, \mathbf{d}) = \sum_{(i,j) \in \mathcal{E}} | \mathbf{r}_{vis}^{ij} |^2_{\Sigma_{ij}} \\ + &\lambda_{odo} \sum_{(i,j) \in \mathcal{E}} | \text{Trans}(\mathbf{G}_{ij}) - \Delta \mathbf{O}{ij} |^2_{\Sigma^o_{ij}}
\end{aligned}
\end{equation}
where $\mathbf{G}_{ij} = \mathbf{G}_j \mathbf{G}i^{-1}$ represents the relative SE(3) transformation, and $\Delta \mathbf{O}{ij}$ is the metric translation constraint from the robot's proprioception.

\subsection{Perturbation Analysis}
To minimize the energy, we apply a small perturbation $\boldsymbol{\xi} \in \mathfrak{se}(3)$ to the poses. The perturbed relative transformation $\mathbf{G}_{ij}'$ can be expressed as:

$$
\begin{aligned}\mathbf{G}_{ij}' &= \exp(\boldsymbol{\xi}_j) \mathbf{G}_j (\exp(\boldsymbol{\xi}_i) \mathbf{G}_i)^{-1} \\
&= \exp(\boldsymbol{\xi}_j) \mathbf{G}_j \mathbf{G}i^{-1} \exp(-\boldsymbol{\xi}_i) \\
&= \exp(\boldsymbol{\xi}_j) \mathbf{G}_{ij} \exp(-\boldsymbol{\xi}_i)
\end{aligned}
$$

Using the adjoint property of the SE(3) group, $\mathbf{G} \exp(\boldsymbol{\xi}) = \exp(\text{Adj}{\mathbf{G}} \boldsymbol{\xi}) \mathbf{G}$, we rewrite the expression as:
\begin{equation}
\mathbf{G}_{ij}' = \exp(\boldsymbol{\xi}_j) \exp(-\text{Adj}{\mathbf{G}_{ij}} \boldsymbol{\xi}_i) \mathbf{G}_{ij}
\end{equation}
Applying the first-order Taylor expansion $\exp(\boldsymbol{\xi}) \approx \mathbf{I} + \boldsymbol{\xi}^\wedge$, we obtain the linearized perturbation:
\begin{equation}
\mathbf{G}_{ij}' \approx (\mathbf{I} + \boldsymbol{\xi}_j^\wedge - (\text{Adj}{\mathbf{G}_{ij}} \boldsymbol{\xi}_i)^\wedge) \mathbf{G}_{ij}\end{equation}

\subsection{Jacobian of the Odometry Residual}
The odometry residual is defined as $\mathbf{r}^o_{ij} = \text{Trans}(\mathbf{G}_{ij}) - \Delta \mathbf{O}_{ij}$. The derivative of the translation component with respect to the Lie algebra perturbation $\boldsymbol{\xi} = [\boldsymbol{\rho}, \boldsymbol{\phi}]^\top$ is given by:
\begin{equation}
\frac{\partial \text{Trans}(\mathbf{G}_{ij})}{\partial \boldsymbol{\xi}_j} = \mathbf{J}^o_j = [\mathbf{I}_{3 \times 3} \mid -[\mathbf{t}_{ij}]_\times] \in \mathbb{R}^{3 \times 6}
\end{equation}
By applying the chain rule and the adjoint property derived in the previous section, the Jacobian with respect to the starting frame $i$ is:
\begin{equation}
\frac{\partial \text{Trans}(\mathbf{G}_{ij})}{\partial \boldsymbol{\xi}_i} = \mathbf{J}^o_i = -[\mathbf{I}_{3 \times 3} \mid -[\mathbf{t}_{ij}]_\times] \cdot \text{Adj}{\mathbf{G}_{ij}}\end{equation}Thus, the linearized odometry residual $\Delta \mathbf{r}^o_{ij}$ is approximated as:\begin{equation}\Delta \mathbf{r}^o_{ij} \approx \mathbf{J}^o_j \Delta \boldsymbol{\xi}_j + \mathbf{J}^o_i \Delta \boldsymbol{\xi}_i\end{equation}

\subsection{Normal Equations}
By accumulating the contributions from both visual and odometry terms, we construct the Schur complement system. The resulting normal equations are:

\begin{equation}
\begin{bmatrix}\mathbf{B}_{vis} + \mathbf{B}_{odo} & \mathbf{E} \\ \mathbf{E}^\top & \mathbf{C} + \lambda \mathbf{I}\end{bmatrix}\begin{bmatrix} \Delta \boldsymbol{\xi} \\ \ \Delta \mathbf{d} \end{bmatrix} =\begin{bmatrix} \mathbf{v}_{vis} + \mathbf{v}_{odo} \\ \mathbf{w} \end{bmatrix}
\end{equation}where $\mathbf{B}_{odo}$ and $\mathbf{v}_{odo}$ represent the Hessian and gradient contributions from the metric anchors, respectively. Specifically, for each edge $(i,j)$, the odometry update to the diagonal blocks of the Hessian is computed as:

\begin{equation}\mathbf{B}_{odo}^{(ii)} = \sum_{j} (\mathbf{J}^o_i)^\top (\Sigma^o_{ij})^{-1} \mathbf{J}^o_i, \quad \mathbf{B}_{odo}^{(jj)} = \sum_{i} (\mathbf{J}^o_j)^\top (\Sigma^o_{ij})^{-1} \mathbf{J}^o_j
\end{equation}


\end{document}